\documentclass[11pt,a4paper]{article}
\usepackage[hyperref]{naaclhlt2019}
\usepackage{times}
\usepackage{latexsym}

\usepackage{url}

\usepackage{times}
\usepackage{helvet}
\usepackage{courier}
\usepackage{url}  
\usepackage{graphicx}  

\usepackage{mdframed}
\usepackage{colortbl}
\usepackage{multirow}
\usepackage{subfigure}
\usepackage{latexsym}
\usepackage{todonotes}
\usepackage[english]{babel}
\usepackage{blindtext}
\usepackage{amsmath}
\usepackage{bm} 
\usepackage{caption}
\usepackage[inline,shortlabels]{enumitem}
\usepackage{color}
\usepackage{todonotes}
\usepackage{subfigure}
\usepackage{float} 
\usepackage{float} 

\DeclareMathOperator{\softmax}{softmax}
\DeclareMathOperator{\concat}{concat}
\DeclareMathOperator{\cosine}{cosine}
\DeclareMathOperator{\average}{average}
\DeclareMathOperator{\Gumbel}{Gumbel}

\newcommand{\SYSNAME}{SURFACE}
\newcommand{\supported}{\textsc{supported}}
\newcommand{\refuted}{\textsc{refuted}}
\newcommand{\unsure}{\textsc{unsure}}

\newcommand{\verifier}{\textsc{Verifier}}
\newcommand{\verifierMT}{\textsc{Verifier + MultiTask}}
\newcommand{\verifierG}{\textsc{Verifier + MultiTask using Gumbel Utility}}

\newcommand{\comment}[1]{}

\newcommand{\smallermath}{\fontsize{7pt}{8.04pt}}
\newcommand{\grayrow}{\rowcolor[gray]{.95}}

\newlength\Fcolumnseprule
\makeatletter
\newcommand\ShowInterColumnFrame{
	\patchcmd{\@outputdblcol}
	{{\normalcolor\vrule \@width\columnseprule}}
	{\vrule \@width\Fcolumnseprule\hfil
		{\normalcolor\vrule \@width\columnseprule}
		\hfil\vrule \@width\Fcolumnseprule
	}
	{}
	{}
}

\makeatother

\usepackage{cleveref}
\crefname{section}{s}{ss}
\Crefname{section}{s}{ss}
\crefname{table}{Table}{}
\crefname{figure}{Fig.}{}
\crefname{algorithm}{Alg.}{}
\crefname{equation}{Eq.}{}
\crefname{appendix}{Appendix}{}
\crefformat{section}{\S#2#1#3}  

\aclfinalcopy 

\title{SURFACE: Semantically Rich Fact Validation with Explanations}

\author{Ankur Padia, Francis Ferraro, and Tim Finin \\
  Univeristy of Maryland, Baltimore County, USA \\
  {\tt \{pankur1, frank, tim\}@umbc.edu} 
}

\date{}

\begin{document}
\maketitle
\begin{abstract}
Judging the veracity of a sentence making one or more claims is an important and challenging problem with many dimensions.  The recent FEVER task asked participants to classify input sentences as either \supported{}, \refuted{} or \unsure{} using Wikipedia as a source of true facts.  \SYSNAME{} does this task and explains its decision through a selection of sentences from the trusted source.  Our multi-task neural approach uses semantic lexical frames from FrameNet to jointly (i) find relevant evidential sentences in the trusted source and (ii) use them to classify the input sentence's veracity.  An evaluation of our efficient three-parameter model on the FEVER dataset showed an improvement of 90\% over the state-of-the-art baseline on retrieving relevant sentences and a \~70\% relative improvement in classification.


\end{abstract}

\section{Introduction}

\noindent
The Web and social media have made it easier for people and organization to share information.  Unfortunately, the information is often false, whether the source knows it or not. Even worse, a recent study \cite{vosoughi2018spread} showed that false information spreads online ``significantly farther, faster, deeper, and more broadly than the truth'' in all of the categories of information studied.  This phenomenon has raised interest in developing automated techniques that can help combat the problem by analyzing a natural language sentence making one or more claims and classifying its as likely to be true, likely to be false or uncertain.

In general, this is a hard task for humans and even more so for machines. One reason is that there are many reasons to consider a fact to be suspicious.  For example, the source (if known) may be untrustworthy or have a motivation to lie, the information may be dated, it might be in conflict with common-sense knowledge, or it might simply reflect an falsehood that is commonly believed by many.  Computer systems that extract facts from text suffer from the additional problem of faulty algorithms that misunderstand the meaning of the text, introducing incorrect statements in their knowledge graphs.

We describe \SYSNAME{}, a system that takes an English sentence that makes one or more claims and classifies it as either as either \supported{}, \refuted{}, or \unsure{} using the text in Wikipedia as a reasonably trustworthy source of true facts.  It also selects a set of Wikipedia sentences to support, justify, and explain its decision.


Verifying the information from the textual claim relies on two broad and complex subtasks: the first is an information retrieval based task to help identify pieces of evidence that could help support or refute the claim. %
The second subtask is classifying the claim, based on the retrieved evidence. To understand better, consider \cref{tab:example}. For the given claim, relevant sentences are required to be identified from the entire corpus, like Wikipedia. Such a searching process results in finding relevant sentences, among which some are actually helpful for verification, some less so, and others not at all. In this example, three evidence sentences are retrieved out of which Evidence$_ 1$ and Evidence$_ 2$ are clearly useful while Evidence$_3$ is less so. Therefore, the full claim verification process can be broken down into four individual steps:
\begin{enumerate*}[(1)]
\item identifying relevant documents, %
\item extracting relevant sentences from them, %
\item determining each sentence's utility for the claim verification, and %
\item classifying the claim, based on the retrieved evidence. %
\end{enumerate*}



\begin{table}
	\small \centering
	\renewcommand{\arraystretch}{1.2}
	\begin{tabular}{|p{.93\columnwidth}|} \hline
		\textbf{Claim}: \textit{Maximum Overdrive} is only a 1980 romance film.\\ \hline
		\textbf{Output}: \refuted{} \\ \hline
		\textbf{Evidence-1 $(\bm{e_1})$}: Maximum Overdrive is a 1986 American science fiction \underline{horror dark comedy} film written and directed by Stephen King.\\ 
		\textbf{Evidence-2 $(\bm{e_2})$}:In 1988 , Maximum Overdrive was nominated for `` Best Film '' at the International Fantasy Film Awards .\\
		\textbf{Evidence-3 $(\bm{e_3})$}: The film stars Emilio Estevez , Pat Hingle , Laura Harrington , and Yeardley Smith.\\ \hline
	\end{tabular}
	\caption{\label{tab:example} \SYSNAME \ takes a sentence making a claim, classifies it as  \supported{}, \refuted{}, or \unsure{} and provides evidence explaining the decision in the form of sentences extracted from a trusted text corpus.}
\end{table}

\SYSNAME{} addresses the problem of a three-label classification of the claim---as either \supported{}, \refuted{} or \unsure{}---and provides an explanation in the form of evidential sentences that were most useful in making the decision. There have been many approaches to identify ``fake news,'' (see our discussion later in the paper) but most have focused on the classification itself and few on explaining why the classification was made \cite{thorne2018fever}. %

In this paper we demonstrate and discuss how to use a semantic frame-based approach for claim verification. We describe an end-to-end system that %
\begin{enumerate*}[(i)]
\item learns the sentence representation from the frame-based evidence sentences, %
\item determines their utility, and %
\item classifies a claim either as \supported{}, \refuted{} or \unsure{}. %
\end{enumerate*} %
We present three parameter-efficient end-to-end models with increasing levels of sophistication: \verifier, \verifierMT, and \verifierG. Multi-tasks model performs better than the recent state-of-the-art approach on the large FEVER dataset that contains nearly 180,000 claims extracted from Wikipedia and annotated by a team of researchers.

\begin{figure*}[h!]
	\centering 
	\vspace{0.1cm}
	\includegraphics[scale=0.43]{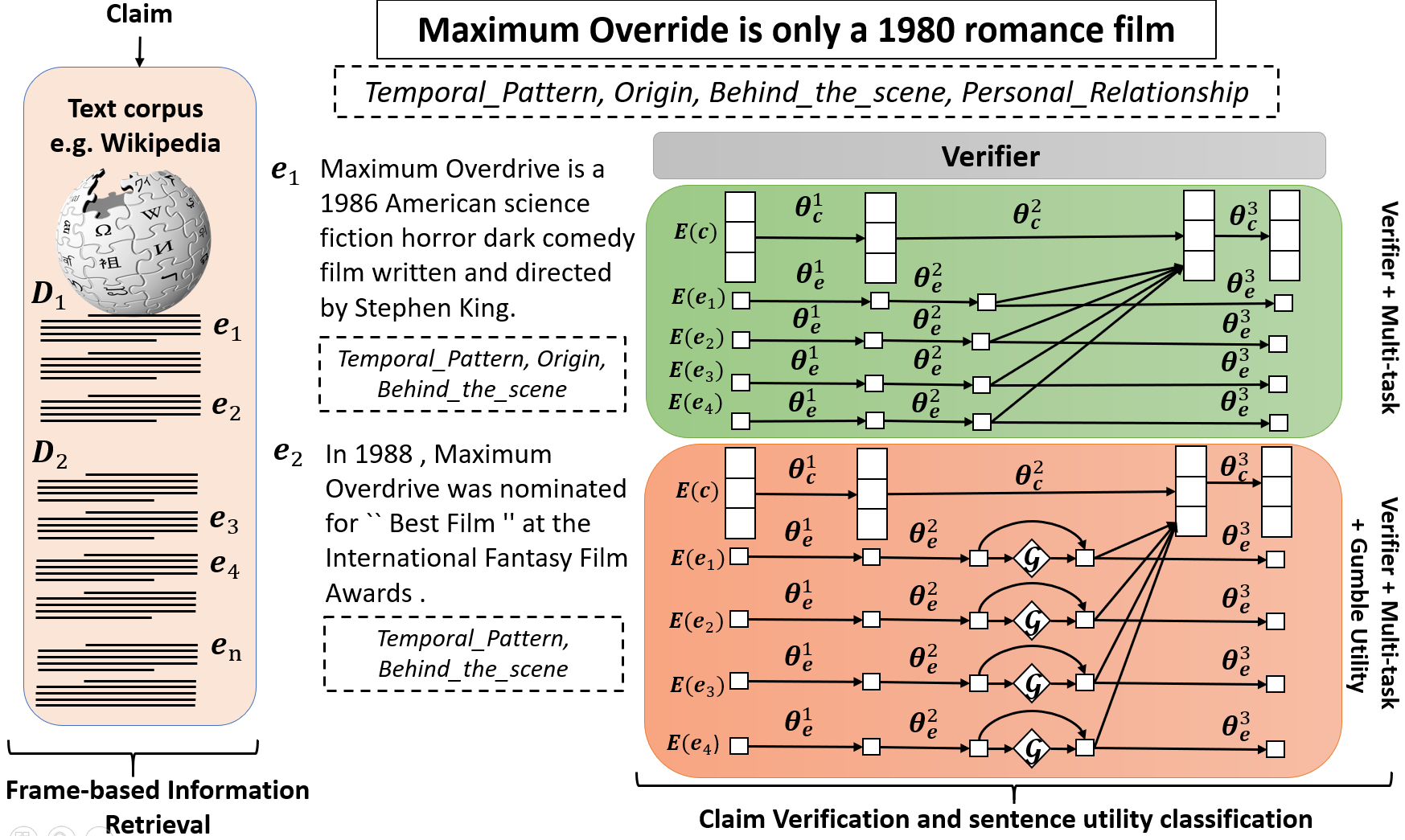}
	\caption{\label{fig:models} Architectural diagram of \SYSNAME{}'s approach with three parameter-efficient models with increasing level of diversity using semantically rich information retrieval. Here $e_1$ and $e_2$ are sample evidence sentences. Dashed box immediate below the text indicates the frames triggered by the text.}
\end{figure*}

\section{Related Work}

Although there has been considerable work fact and news verification, no work has explored the use of semantically rich lexical resources like FrameNet in the process. We briefly review related work on two variations on the problem: verifying facts from a knowledge graph, such as Wikidata, and verifying claims directly from text, such as sentences in a newswire article.

\subsection{Schematic Fact Verification}

We describe approaches that assume the claim to be verified is expressed in a structured format, typically as a triple in a knowledge graph expressing a relation between a subject and object, e.g. (:BarackObama, :president\_of, :Germany), grounded in a schema (minimal or rich).

Web search approaches, like OpenEval \cite{samadi2013openeval}, analyze search engine results to verify if the fact is correct or incorrect.  Simple systems do not consider the bias or reliability of retrieved text sources, a drawback addressed by \citet{samadi2016claimeval}, which considers the contradictory information from sources and use it to assign score to each of the fact using probabilistic soft logic \cite{brocheler2012probabilistic}. 

Rule-based systems use a fixed set of rules, typically grounded in logic, and formatted as an optimization problem. \citet{pujara2013knowledge} scores a claim using rules that check schema constraints (e.g., a relation's domain, range, cardinality) and plausible inferences involving other claims using Markov Logic Network \cite{richardson2006markov}, projecting the problem into a continuous space and scoring facts on how well they satisfy the rules. The approach was extended to assign scores to sets of claims, rather than individual ones, with minimal supervised data \cite{KGEval}.  Such rule-based approaches are limited by the richness of the knowledge graph's schema and additional inference rules, both of which require manual construction or an independent (and difficult) learning task.

Some systems use an an ensemble-based approach where a claim's authenticity is determined using multiple variations of the claims.
\citet{yu2014wisdom} uses an unsupervised iterative approach to distill truthful from false facts and \citet{nakashole2014language} determines the truthfulness of the fact using linguistic features and evaluates the objectivity of the documents. Ensemble approaches are helpful but their reliance on diverse variants is a serious limiting factor. 

\SYSNAME{}'s approach avoids these limitations as it is not dependent on black-box search engines, does not use low-coverage and hard-to-maintain rules, and doesn't require multiple claim instances. Its use of \textit{frame}-analyses for both claims and evidence significantly improve matching. \citet{padia2018kgcleaner} use frame-analysis but restrict their experiments to schema-backed claims, and do not consider free text claims. However, they restrict their experiments to schema-backed claims, and they do not consider free text, or non-schematic, claims as we do here.

\subsection{Text Based Claim/News Verification}


Most work on evaluating facts in textual sources like newswire and social media uses supervised approaches and manual feature engineering. \citet{fncBaseline} uses word n-grams and sets of predefined refuting words. \citet{perez2017automatic} builds an SVM classifier that uses n-grams, psycholinguistic features, syntactic, and other linguistic features 
to identify whether an entire news article is suspect. Such approaches requires expert knowledge to manually design features for the task. The limitations of feature engineering was relaxed and, in some cases, performance improved by using neural network based approaches to learn a claim and evidence representation automatically from the text. \citet{zengNeuralStance} show that bi-directional RNN models outperform hand-crafted features. Additionally, \citet{talosFNC} demonstrate that combining features and representation of the claim and evidence helps to achieve better performance. One of the winning teams of the Fake News Challenge \cite{fnc} used a single layer perceptron with term frequency as the input vector along with the cosine similarity score between the claim and evidence for classification \cite{riedel2017simple}.

\citet{kochkina2018all} uses a multi-task neural network to learn representations for stance and  rumor detection, tracking, and verification. Our work differs by exploring the impact of semantically rich, frame-based evidence retrieval and matching and using a multi-task approach to learn better representations for classification and selecting evidence to explain the classification.



Even though several studies have been conducted to verify the claim of different types, no extensive study is made to understand the combination of semantically-informed feature with recent neural approaches for claim verification. While \citet{padiaFever2018} use semantic parses to assist claim verification, their study is limited: they do not study how evidence and claim representation affect claim verification performance. In this study we fill this gap by proposing multiple models to understand the effect of sentence representation for claim verification task and also provide insights into when one model is better then another.

\section{Approach}
\label{sec:approach}

We describe each of the steps of our approach to verify the truthfulness of a \textit{claim}. Input to the system is the text of a claim and the output is a classification of the claim as \supported{} or \refuted{} or \unsure{}. Additionally, if the decision is \supported{} or \refuted{}, an evidence sentence is also provided to explain the decision.  For each input claim the following three steps are performed: %
\begin{enumerate*}[(1)]
\item \textit{evidence retrieval} finds potentially useful sentences to either help support or refute the claim; %
\item \textit{evidence utility} (optionally) learns to predict how useful each piece of retrieved evidence is; and %
\item \textit{verification} or \textit{refutation} of the claim, based on the evidence and its predicted utility. %
\end{enumerate*}
\Cref{fig:models} summarizes the process and architecture of each of our three models---(i) \verifier, (ii) \verifierMT, and (iii) \verifierG.

We process each of the claims using a named entity recognition system to identify the named entities and \textit{frame extraction} to annotate the claim with frames from FrameNet 1.7, using the the Thrift-based \textsc{Concrete} schema for annotation representation. Similarly, we used a previously annotated version of Wikipedia in which each of the documents were annotated with same named entity tagger and frame annotator \cite{ferraro2014concretely}.

\subsection{Evidence Retrieval} %

We decompose the evidence retrieval task into two separate subtasks: \textit{document retrieval} finds relevant documents that contain possible evidence and \textit{sentence retrieval} to select the relevant evidence sentences from the retrieved documents.  We note that this decomposition is a heuristic approach for evidence retrieval; theoretically, our models are not restricted to this two-step approach.

\subsubsection{Document Retrieval.}
To find potentially relevant semantically annotated Wikipedia articles, we first identified entities mentioned within the claim. %
Then, we attempted to match each entity to a corresponding Wikipedia page. %
We used an exact string match to associate entities with a Wikipedia page. %
As our aim in this work is to examine semantic retrieval and classification, we intentionally opted for a straight-forward initial retrieval step; while undoubtedly a more complex document retrieval could result in an improved classification system, our results demonstrate that even simple lexical matching methods can not only be effective, but can lead to a 100\% improvement over prior efforts. %

\subsubsection{Sentence Retrieval.}
For each of the retrieved documents, we select sentences that contain at least one of the frames that was triggered by the claim using exactly matches. %
We can supplement the semantic retrieval with task-specific retrieval information. %
Consider that we might know a priori that evidence is likely to come from certain parts of a document. For example, as with \citet{thorne2018fever}'s FEVER task, evidence sentences are almost all drawn from the first few sentences of a document. %
Knowing information like this about a particular task can help ensure that the system can both learn to recognize and utilize relevant, important pieces of evidence, and also learn to discount irrelevant or bad pieces of evidence. %
That is, while we do not want to blindly rely, \textit{prima facie}, on all possible task-specific sentences, retrieving and then using them wisely can help the overall model focus on learning effective classification representations. %

We note that further post-processing on the retrieved sentences can be done, depending on context and task specifications.
We consider further processing of the evidence in the multi-task setting (explained later) depending on the sentence representation.

\subsection{Evidence Utility Prediction.}
The retrieval described above---using only coarse named entity and semantic frame matching---is a general procedure and does not utilize any task or corpus-specific tailoring. %
To hepl refine the retrieval, our models allow each piece of evidence to be evaluated for its utility in the task at hand. %
Though the precise formulation varies based on the model, at a high-level this utility prediction is simply a binary classifier, e.g., logistic regression, that uses the learned sentence representations $e_i^{(2)}$ to predict the utility $\hat{u_i} \in \{0,1\}$, e.g., with a logistic regression model $p(\hat{u_i}\ |\ e_i^{(2)})$. %

While the utility predictor can help refine the retrieval to learn what are truly uninformative sentences (e.g., those sentences that were erroneously retrieved), it can also learn task-dependent specifications on what should or should not be retrieved. %
For example, a retrieved evidence sentence may not have anything to do with the claim; in this case, it would have been retrieved erroneously. %
On the other hand, an evidence sentence may be topically relevant, but, for idiosyncratic reasons, it is deemed to be irrelevant.\footnote{ %
   These concerns are not theoretical; previous well-known information extraction tasks encounter these types of relevancy issues \cite{sundheim1992muc4,duc2004}. %
} %


\subsection{Claim Classification.}
In this section we describe our 3 models  (i) \verifier, (ii) \verifierMT, and (iii) \verifierG . 
Each of the retrieved sentences in the previous step could be in the following category (1) the sentence was useful and helps in making decision about claim truthfulness, (2) the sentence was useful but was outside the pre-decided scope within the document document (due to a design choice), and (3) the sentence can be not useful irrespective of the pre-decided scope of the evidence.

\subsubsection{Claim and Evidence representation.} We consider the average bag-of-word embeddings for a given claim and evidence as its initial representation. We used pretrained word embeddings from Glove \cite{pennington2014glove} to compute the average of the word embeddings. We associate the claim with the evidence using cosine similarity. As described in our experimental section, initializing each claim's and evidence's representation with their averaging bag-of-words embeddings resulted in significant performance gains against the previous FEVER baseline approaches \cite{thorne2018fever}. Although in many of our experiments we refine these initial sentence representations with MLPs, as we discuss, more diverse and sophisticated techniques can be appliedwhich will most likely improve the results. %


\subsubsection{Mapping out-of-scope frame-based evidence to in-scope evidence.}
Scope of the evidence for a claim plays an important role in determining the truthfulness of the claim. Assuming the scope is limited to introduction of a Wikipedia page, a mapping technique is required to map the frames based evidence sentences (especially out of scope of the introduction but could be still relevant). We use Jaccard token based similarity between the introduction sentences and the frame sentences and use Hungarian algorithm \cite{kuhn55} to find optimal mapping. %
Other similarity measures beside Jaccard, like neural-based sentence similarity or other classifier-based similarity techniques, could be used to match sentences. Both due to space constraints and overall ease of understanding, we use the Jaccard similarity---a widely applicable and well-studied measure---exclusively in our experiments. %

\subsubsection{Model Specifications.}
Our three models, (i) \verifier{} , (ii) \verifierMT{} (Verifier+MT), and (iii)\verifierG{} (Verifier+MT+G), 
all models take claims and retrieved evidence sentences as input. Both Verifier+MT and Verifier+MT+G use a multi-task setting to verify the claim and judge each evidence's utility; the differences between Verifier+MT and Verifier+MT+G are in how they use these utility judgments to further fine tune each sentence's representation. %
Unlike both Verifier+MT and Verifier+MT+G, Verifier classifies only the claim, using all retrieved evidence and without learning a utility model over the evidence. 

\paragraph{Model 1: \verifier{}.}
We set \verifier to a parameter-efficient two-layer tron MLP.
Figure \ref{fig:models} shows that each layer in the model learn a more abstract representation than the previous layer. Eq.(\ref{eqn:v1}-\ref{eqn:v4}) describes the process. 
{\smallermath
  \begin{eqnarray} 
	h^{(0)} & = & \concat\left(\overrightarrow{c},\overrightarrow{e},\cosine\left(\overrightarrow{c},\overrightarrow{e}\right)\right)\label{eqn:v1}\\
	h^{(1)} & = & g^{(1)}\left(W^{(1)}h^{(0)}+b^{(1)}\right)\label{eqn:v2}\\
	h^{(2)} & = & g^{(2)}\left(W^{(2)}h^{(1)}+b^{(2)}\right)\label{eqn:v3}\\
	\hat{y} & = & \softmax\left(W^{(3)}h^{(2)}+b^{(3)}\right)\label{eqn:v4}
\end{eqnarray}}
\noindent The activation functions $g^{(1)}, g^{(2)}$ are set to rectified linear unit (ReLU) \cite{krizhevsky2012imagenet} and $\softmax$ is provides the probability distribution over the labels. We concatenate the evidence sentences and combine the average of word embeddings from the claim, average of word embeddings for the evidence sentences, and the cosine similarity between the claim and the averaged evidence. 

\paragraph{Model 2: \verifierMT{}.}
We consider the \verifier from the previous section for multi-task capability to produce \verifierMT. The \verifier{} provides the classification from the claim but does not provide insight about the \textit{utility} evidence sentence as the evidence frame based sentences could be useful or not useful depending on the scope of the evidence and relevance with the claim. Hence we consider multi-task setting for the model where the first-task is to identify the truthfulness of the claim (3-way classification), and the second task is to identify the utility of the evidence sentences (2-way classification) if a sentence is useful or not.

Consider the equations Eq. \ref{eqn:mtm1} - \ref{eqn:mtm9} into three parts, learning the claim representation (Eq.  \ref{eqn:mtm1} - \ref{eqn:mtm3} ), learning evidence representation (Eq.  \ref{eqn:mtm4} - \ref{eqn:mtm7}), and classification (Eq  \ref{eqn:mtm8} - \ref{eqn:mtm9}).  

Here $E(\cdot)$ is a embedding function that provides the initial representation for sentence and claim . Here  $c^{(1)}, c^{(2)}$ is the abstract representation of the claim. We set activation function $g_{c}^{(1)}$ and $g_{c}^{(2)}$ to be ReLU. More abstract representation is obtained using \eqref{eqn:mtm2} and \eqref{eqn:mtm3}.
{\smallermath
 \begin{eqnarray}
    c^{(0)} & = & E\left(c\right) \label{eqn:mtm1}\\
    c^{(1)} & = & g_{c}^{(1)}\left(W_{c}^{(1)}c^{(0)}+b_{c}^{(1)}\right)\label{eqn:mtm2}\\
    c^{(2)} & = & g_{c}^{(2)}\left(W_{c}^{(2)}c^{(1)}+b_{c}^{(2)}\right)\label{eqn:mtm3}
 \end{eqnarray}
}
Similarly, we learn evidence representation and compute the cosine
similarity to associate it with the claim. We sample $K$ evidence from
the pool of evidence for a given claim. We learn common representation
among the sentences by sharing the parameters $ W_{e}^{(1)},
W_{e}^{(2)}, W_{e}^{(3)}, b_{e}^{(1)}, b_{e}^{(2)}, b_{e}^{(3)} $
among the sampled sentences with the intuition that the words used
across the sentences will provide useful clue to the classifiers and
help differentiate useful sentences from the not useful
ones. Moreover, sharing the parameter help reduces the number of
parameters\footnote{We tried separate parameters for each
evidence and found performance to be poorer compared to shared parameter models. }. Once latent representation is learned we take the average $e_{avg}$ across all the evidence representation to be considered for final classification. 
{\smallermath
  \begin{eqnarray}
    e_{i}^{(0)} & = & \concat\left(E\left(e_{i}\right),\cosine\left(c^{(0)},E\left(e_{i}\right)\right)\right)\label{eqn:mtm4}\\
    e_{i}^{(1)} & = & g_{e}^{(1)}\left(W_{e}^{(1)}e_{i}^{(0)}+b_{e}^{(1)}\right)\label{eqn:mtm5}\\
    e_{i}^{(2)} & = & g_{e}^{(2)}\left(W_{e}^{(2)}e_{i}^{(1)}+b_{e}^{(2)}\right)\label{eqn:mtm6}\\
    e_{avg} & = & \underset{i}{\average}\left(e_{i}^{(2)}\right)\label{eqn:mtm7}
  \end{eqnarray}
}
Finally, we concatenate the evidence representation and the claim representation and find a probability distribution across all claim labels using $\softmax$ (Eq. \ref{eqn:mtm8}). Similarly, we find sentence utility for each of the $K$ sentence (Eq \ref{eqn:mtm9}).
{\smallermath
  \begin{eqnarray}
    \hat{y_{c}} & = & \softmax\left(W_{c}^{(3)}\concat\left(c^{(2)},e_{avg}\right)+b_{c}^{(3)}\right)\label{eqn:mtm8}\\
    \hat{u_{i}} & = & \softmax(W_{i}^{(3)}e_{i}^{(2)}+b_{e}^{(3)})\label{eqn:mtm9}
  \end{eqnarray}
}

\paragraph{Model 3: \verifierG{}.} Computing evidence sentence utility in the multi-task setting gives
insight about the evidence, but the model is unable to modify the sentence representation with its utility for classification because the values are discrete rather than continuous.  Using the utilities directly in the classification of the claim in the last layer of the model is not possible as the stochastic neural networks cannot update its weights in backpropagation. We approximate the discrete utility variable with a continuous variable using Gumbel-Softmax trick \cite{jang2016categorical}, yielding the \verifierG{} model to incorporate utilities of the sentences at the time of predicting claim classification.
{\smallermath
	\begin{eqnarray}
	c^{0} & = & E\left(c\right)\label{eqn:g1}\\
	c^{1} & = & g_{c}^{1}\left(W_{c}^{1}c^{0}+b_{c}^{1}\right)\label{eqn:g2}\\
	c^{2} & = & g_{c}^{2}\left(W_{c}^{2}c^{1}+b_{c}^{2}\right)\label{eqn:g3}\\
	e_{i}^{0} & = & \concat\left(E\left(e_{i}\right),\cosine\left(c^{0},E\left(e_{i}\right)\right)\right)\label{eqn:g4}\\
	e_{i}^{1} & = & g_{e}^{1}\left(W_{e}^{1}e_{i}^{0}+b_{e}^{1}\right)\label{eqn:g5}\\
	e_{i}^{2} & = & g_{e}^{2}\left(W_{e}^{2}e_{i}^{1}+b_{e}^{2}\right)\label{eqn:g6}\\
	z_{i} & = & \Gumbel\left(W_{e}^{3}e_{i}^{2}+b_{e}^{3}\right)\label{eqn:g7}\\
	o_{i} & = & e_{i}^{2}z_{i}^{T}\label{eqn:g8}\\
	R & = & \underset{i}{\average}\left(o_{i}\right)\label{eqn:g9}\\
	\hat{y_{c}} & = & \softmax\left(W_{c}^{3}\concat\left(c^{2},R\right)+b_{c}^{3}\right)\label{eqn:g10}\\
	\hat{u_{i}} & = & \softmax(W_{i}^{4}o_{i}+b_{e}^{4})\label{eqn:g11}
	\end{eqnarray}
}
Motivation for the Gumbel-Softmax distribution is Gumbel-Max \cite{gumbel1954statistical,maddision2014gumbletrick} trick where the idea is to add Gumbel noise with size same as the number of discrete variable independently sampled and taking the \textit{argmax} hence the name Gumbel-Max. However, the \textit{argmax} is not a continuous operation and hence \textit{argmax} is approximated with Softmax to make it continuous. The quality of the approximation of the Gumbel-Softmax distribution from a discrete distribution is handled by a hyper parameter called temperature $\tau$. Lower the value of $\tau$ closer the distribution to discrete distribution. We consider the binary utility classification of the sentence as discrete and add Gumbel noise to it. Broader understanding about the topic can be found in \cite{jang2016categorical}.

For this model we keep the procedure to learn the claim representation to be the same as (Eq. \ref{eqn:mtm1}-\ref{eqn:mtm3}). However, we modify the  operation to classify the utility of sentences using (Eq.\ref{eqn:g1}-\ref{eqn:g11})
The difference between the \verifierG{} and \verifierMT{} is consideration of sentence representation. In \verifierG{} the abstract sentence representation is modified with the Gumbel utility with the outer product, and hence the model considered modified sentence representation.

\section{Experiments}
\label{sec:experiments}
In this section we explore and describe the impact of frame-based semantic retrieval on classifier ability to discriminate false claims from true ones. We consider the following research questions:
\begin{enumerate*}[(i)]
\item How good is IR performance using only frame-based sentence retrieval? 
\item How does the multi-tasking of utility and claim classification effect the overall performance? 
\item Does changing the number of evidence sentences influence performance? 
\item Does combining frame sentences with FEVER sentences with limited scope effect performance? 
\end{enumerate*}




\subsubsection{Datasets.}
We used the large FEVER dataset with 145.5K \textit{Train} and 20K \textit{Development} claims along with human-selected evidence sentence from Wikipedia article introduction sections. Since the dataset's 20K test claims have not been released, we evaluate our approach on the development dataset and on the training dataset with five-fold cross validation. Statistics for each are shown in Table \ref{tab:stats}.
We used two sets of sentences: FEVER sentences drawn from the introduction section of the Wikipedia articles, and frame-annotated sentences from the entire Wikipedia article. We collected and trained the model based on the frame based sentences as described in the approach section. 


\subsubsection{Performance Metrics, Baseline and Models.}
We used the FEVER-provided script \cite{FeverScript} that considers a claim correctly classified when both the IR and classification parts are correct. For \supported{} and \refuted{} the predicted evidence set should match the human annotators' and mismatches effect the precision and recall metrics.
For \unsure{}, the label must be correct and evidence criteria is relaxed. It uses precision, recall and F1 for IR and accuracy for classification. It combines IR and classification performance to compute the overall score, called \textit{FEVER score}.\footnote{We used momentum based Stochastic Gradient Descent with learning rate of 0.01, decay of 1e-6, and momentum of 0.9. We set L2 regularizer to 0.1, dropout of 0.5 only for first verifier in our case MLP and trained for 50 epoch.}


\begin{table}
  \centering
  \begin{tabular}{|c|c|c|c|} \hline
       \grayrow
        \textbf{Dataset} & \textbf{\#claims} & \textbf{Avg.}   & \textbf{Avg.} \\
	   \grayrow
        & & \textbf{\#Doc (IR)} & \textbf{\#Sent. (IR)}\\
	\hline
	\textbf{Train} & 145,449 & 18.631 & 1.1348\\
	\hline
	\textbf{Dev} & 19,998 & 9.5828 & 1.1082\\
	\hline
	\end{tabular}
  \caption{\label{tab:stats} Statistics for claims from FEVER dataset (from Wikipedia articles introduction sections) and number of documents and sentence retrieved by \SYSNAME{} (from the entire Wikipedia articles)}
\end{table}


\begin{table}[t]
  \centering
  \small
    \centering
    \begin{tabular}{|l|c|c|c|}
      \hline
      \grayrow
      \textbf{Approach} & \textbf{P} & \textbf{R} & \textbf{F1} \\
      \hline
      \textbf{Verifier-1} & 0.3772 & 0.2604 & 0.3081 \\
      \hline
      \textbf{Verifier-2} & 0.3772 & 0.2604 & 0.3081 \\
      \hline
      \hline
      \textbf{Verifier-1+MT+G} & 0.3772 & 0.2604 & 0.3081 \\
      \textbf{Verifier-2+MT+G} & 0.3772 & 0.2604 & 0.3081  \\      
      \hline
      \textbf{Verifier-1+MT+G}(u=10) & 0.6153 & 0.0902 & 0.1573 \\
      \hline
      \textbf{Verifier-2+MT+G}(u=10) & 0.6249 & 0.0883 & 0.1547  \\
      \hline        
      \textbf{Verifier-1+MT} & 0.3772 & 0.2604 & 0.3081  \\
      \hline
      \textbf{Verifier-2+MT} & 0.3772 & \textbf{0.2604} & \textbf{0.3081}  \\
      \hline
      \textbf{Verifier-1+MT}(u=10) &  0.6257  &   0.0882 &   0.1546 \\
      \hline
      \textbf{Verifier-2+MT}(u=10) & \textbf{0.6257} & 0.0883  & 0.1548 \\
      \hline
    \end{tabular}
    \caption{\label{tab:train_ir} IR performance on Train, cross validation }
  \end{table}

  \begin{table}[t]
    \centering
    \small
    \begin{tabular}{|l|c|c|}
      \hline
      \grayrow
      \textbf{Approach} & \textbf{FEVER Score} & \textbf{ACC}\\
      \hline
      \hline
      \textbf{Verifier-1} & 0.1443 & 0.55024\\
      \hline
      \textbf{Verifier-2} & 0.1709 & 0.5636 \\
      \hline
      \hline
      \textbf{Verifier-1+MT+G}  & 0.2446 & 0.6495 \\
      \hline
      \textbf{Verifier-2+MT+G} & 0.2564 & 0.6629 \\
      \hline
      \textbf{Verifier-1+MT+G}(u=10)  &  0.1554 &  0.6508 \\
      \hline
      \textbf{Verifier-2+MT+G}(u=10) & 0.1535 & 0.6540 \\
      \hline
      \hline
      \textbf{Verifier-1+MT}  & 0.2504 & 0.6539 \\
      \hline
      \textbf{Verifier-2+MT} & \textbf{0.2564} & \textbf{0.6927} \\
      \hline
      \textbf{Verifier-1+MT}(u=10)  & 0.1529  &  0.6679 \\
      \hline
      \textbf{Verifier-2+MT}(u=10) &  0.1773 & 0.6909 \\
      \hline
    \end{tabular}
    \caption{\label{tab:train_classification} Classification performance on Train, cross validation}
  \end{table}

\begin{itemize} 

\item \textbf{FEVER Baseline.} 
At the time of writing the only peer-reviewed publicly available baseline is from \citet{thorne2018fever}. 
This FEVER baseline does the IR task using TF-IDF with a claim to retrieve documents and selects the \textit{top-l} for evidence extraction. Sentence level TF-IDF is then used to find sentences most similar to the claim. It uses a state-of-the-art RTE model, Decomposable Attention (DA) \cite{parikh2016decomposable}, to model the claim and evidence.

\item \textbf{\verifier.} We evaluated our verifier with one (Verifier-1) and two (Verifier-2) layers. During the evaluation we found slight improvement in performance when the additional layer was added.

\item \textbf{\verifierG{} (Verifier+MT+G).} In addition to learning an abstract representation, we add an auxiliary task of identifying the utility of the sentences to handle cases where the frame-based sentences would be out of the scope of the introduction but may be relevant as evidence. For the multi-task settings with Gumbel-Softmax distribution \cite{jang2016categorical}, we consider two models with one layer (Verifier-1+MT+G) and two layers (Verifier-2+MT+G).

\item \textbf{\verifierMT{} (Verifier+MT) } These are similar to (Verifier+MT+G) but instead of considering the outer product of the sentence latent representation and Gumbel-Softmax utility, they take the direct sentence representation without modification of the evidence sentences for claim and sentence utility classification. 

\end{itemize}


\subsection{Results and Discussion}

We did initial experiments on 5 fold cross validation of the training dataset; these results are given in the Tables \ref{tab:train_ir} and \ref{tab:train_classification}.  
We report our evaluation results on development dataset in Tables \ref{tab:dev_ir} and \ref{tab:dev_classification}. %
As of submission, this development dataset is the only publicly available evaluation dataset for FEVER. %
 

\subsubsection{IR and Classification Performance (Train 5-fold).}
Tables \ref{tab:train_ir} and \ref{tab:train_classification} show the model performance using five-fold training with each fold containing nearly 30K claims. It is clear that two layers perform better compared to one layer. Moreover changing the sentence representation with utility value (Verifier-2+MT+G) with Gumbel noise results in lower performance compared to considering raw sentence representation (Verifier-2+MT). Even though the performance of Gumbel is lower due to its complexity we found increasing training epochs to 100 give some performance gain. Moreover, just performing the frame matching to obtain evidence without post-processing to remove zero utility evidences gets the score higher compared to other models that does post-processing on IR sentences (denoted by \textit{u=10}).

\begin{figure}[t!]
	\centering
	\centering     
	\subfigure[Gumbel ACC]{\label{fig:g3by3acc}\includegraphics[width=35.54mm]{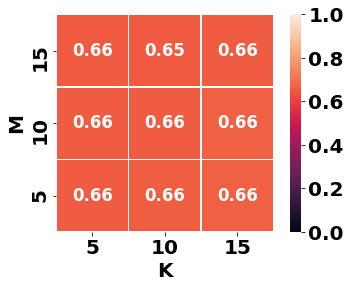}}
	\subfigure[MTM ACC]{\label{fig:m3by3acc}\includegraphics[width=35.5mm]{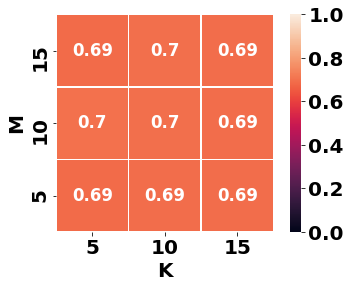}}
	
	\subfigure[Gumbel Precision]{\label{fig:g3by3p}\includegraphics[width=35.54mm]{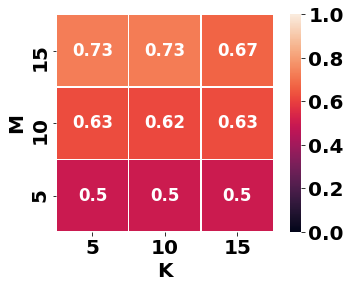}}
	\subfigure[MTM Precision]{\label{fig:m3by3p}\includegraphics[width=35.5mm]{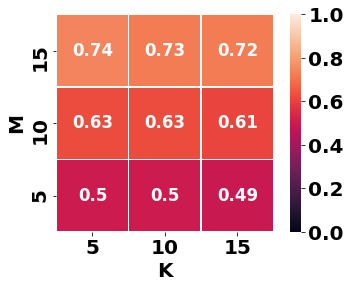}}
	
	
	\subfigure[Gumbel F1]{\label{fig:g3by3f1}\includegraphics[width=35.54mm]{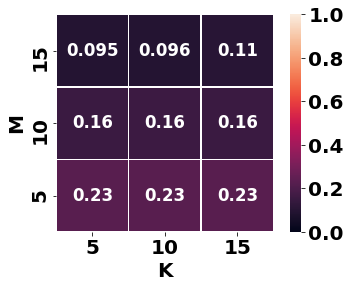}}
	\subfigure[MTM F1]{\label{fig:m3by3f1}\includegraphics[width=35.5mm]{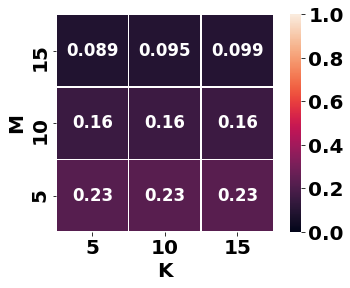}}
	
	\caption{\label{fig:3by3}Performance study of changing K and M values, and using the utilities as post-processing. Darker the color the worser the performance. Heat maps are based on 10 number of sentences. 
	}
        \vspace{-1em}
\end{figure}

\begin{table}[t] \small
   \centering 
   \begin{subtable} \centering
     \begin{tabular}{|l|c|c|c|}
       \hline
       \grayrow
       \textbf{Approach} & \textbf{P} & \textbf{R} & \textbf{F1} \\
       \hline
       \textbf{FEVER (MLP)} & -- & -- & 0.1718 \\
       \textbf{FEVER (DA)} & -- & -- & 0.1718 \\
       \hline
       \hline
       \textbf{Verifier-2} & 0.4524 &  0.2577 &  0.3284 \\
       \hline
       \textbf{Verifier-2+MT+G} & 0.4524 &  0.2577 &  0.3284\\
       \hline
       \textbf{Verifier-2+MT+G}(u=10) & 0.6182 &  0.0897 &  0.1567\\
       \hline
       \textbf{Verifier-2+MT} & 0.4524 &  \textbf{0.2577} &  \textbf{0.3284} \\
       \hline
       \textbf{Verifier-2+MT}(u=10) & \textbf{0.7273} & 0.0533 & 0.0993 \\
       \hline
     \end{tabular}
     \caption{\label{tab:dev_ir} IR Performance on Dev with u=10.}
   \end{subtable}
  ~\begin{subtable} \centering
     \begin{tabular}{|l|c|c|}
       \hline
       \grayrow
       \textbf{Approach} & \textbf{ACC} & \textbf{FEVER Score} \\
       \hline
       \textbf{FEVER (MLP)} &   0.4186 & 0.1942 \\
       \textbf{FEVER (DA)} &   0.5137 & 0.3127 \\
       \hline
       \hline
       \textbf{Verifier-2} & 0.39 & 0.1153  \\
       
       \hline
       \textbf{Verifier-2+MT+G} & 0.6192 & 0.3149 \\
       \hline
       \textbf{Verifier-2+MT+G}(u=10) & 0.6213 & 0.2501 \\
       \hline
       
       \textbf{Verifier-2+MT} & 0.6518 & \textbf{0.3452} \\
       \hline
       \textbf{Verifier-2+MT}(u=10) & \textbf{0.6519} &  0.2554\\
       \hline
     \end{tabular}
     \caption{\label{tab:dev_classification} Classification performance on Dev.}
   \end{subtable}
\end{table}

\subsubsection{IR Performance (Dev).} \label{sec:ir_dev} 
As evident from Table \ref{tab:dev_ir}, our frame-based approach achieves F1 of 0.3284 which is significantly better than the baseline, with a relative improvement of 90\%. 
Moreover, frame-based retrieval gives higher precision and better recall compared to the other non-machine learning based retrieval method like TF-IDF (as used in the baseline models). 
When we consider multitasking (denoted MT), the classification score increases drastically. %
Although we argue that precision is more appropriate for evidence retrieval, we note that post-process removal of evidence can harm recall, and subsequently F1.
 
\subsubsection{Classification Performance (Dev).}  \label{sec:clf_dev}
Table \ref{tab:dev_classification} shows the results on development dataset using FEVER score and the classification accuracy.  Baseline state-of-the-art RTE decomposable attention \cite{parikh16} with TF-IDF retrieval gives better performance compared to \verifier. However, the best classification performance is achieved when the multi-task setting is considered. \verifierMT{} performance is better compared to the rest and Verifier+MT+G model as the Verifier+MT does not change the sentence representation with its outer product of utility and representation. Moreover, considering the post-processing of the retrieved evidences sentences reduces the F1 score eventually impacting the FEVER score and reducing its numbers denoted by \textit{u=10}.

\subsection{Ablation Study}
\label{sec:ablation_study}



We also measured the effect of considering additional sentences from the scope along with the frame sentences. Figures \ref{fig:3by3} show heat maps for varying number of sentences sample from frame sentences (denoted by K) and number of sentences sampled from FEVER corpus (denoted by M). Increasing the number of frame sentences gradually increases the performance and starts to reduce with sentences more than the average number of sentences. Similarly, performance reduces by adding introduction sentences.


\section{Conclusion}
We described \SYSNAME, a prototype automated systems that can test the validity of a sentence making one of more claims and justifies its decision by producing sentences from Wikipedia that support or refute the claims.  Our multi-task neural approach uses semantic lexical frames from FrameNet to jointly (i) find relevant sentences in Wikipedia (ii) use them to classify the input sentence's veracity.  An evaluation of our efficient three-parameter model on the FEVER dataset showed relative improvement of 90\% over the state-of-the-art baseline on retrieving relevant sentences and a nearly 70\% relative improvement in classifying the input sentences.





\fontsize{9.0pt}{10.0pt} 
\bibliography{surface}

\begin{thebibliography}{31}
\expandafter\ifx\csname natexlab\endcsname\relax\def\natexlab#1{#1}\fi

\bibitem[{fnc()}]{fnc}

\newblock {Fake News Challenge}.
\newblock http://fakenews\-challenge\-.org/.

\bibitem[{Brocheler et~al.(2012)Brocheler, Mihalkova, and
  Getoor}]{brocheler2012probabilistic}
Matthias Brocheler, Lilyana Mihalkova, and Lise Getoor. 2012.
\newblock Probabilistic similarity logic.
\newblock \emph{arXiv preprint arXiv:1203.3469}.

\bibitem[{Ferraro et~al.(2014)Ferraro, Thomas, Gormley, Wolfe, Harman, and {Van
  Durme}}]{ferraro2014concretely}
Francis Ferraro, Max Thomas, Matthew~R Gormley, Travis Wolfe, Craig Harman, and
  Benjamin {Van Durme}. 2014.
\newblock Concretely annotated corpora.
\newblock In \emph{AKBC Workshop at NIPS}.

\bibitem[{{FEVER}(2018)}]{FeverScript}
{FEVER}. 2018.
\newblock {FEVER} evaluation script.
\newblock https://github.\-com/\-sheffieldnlp/fever-baselines.

\bibitem[{Galbraith et~al.(2018)Galbraith, Iqbal, van Veen, Rao, Thorne, and
  Pan}]{fncBaseline}
Byron Galbraith, Humza Iqbal, HJ~van Veen, Delip Rao, James Thorne, and Yuxi
  Pan. 2018.
\newblock Baseline fnc implementation.

\bibitem[{Gumbel(1954)}]{gumbel1954statistical}
Emil~Julius Gumbel. 1954.
\newblock \emph{Statistical theory of extreme values and some practical
  applications: a series of lectures}.
\newblock 33. US Govt. Print. Office.

\bibitem[{Jang et~al.(2016)Jang, Gu, and Poole}]{jang2016categorical}
Eric Jang, Shixiang Gu, and Ben Poole. 2016.
\newblock Categorical reparameterization with gumbel-softmax.
\newblock \emph{arXiv preprint arXiv:1611.01144}.

\bibitem[{Kochkina et~al.(2018)Kochkina, Liakata, and
  Zubiaga}]{kochkina2018all}
Elena Kochkina, Maria Liakata, and Arkaitz Zubiaga. 2018.
\newblock All-in-one: Multi-task learning for rumour verification.
\newblock \emph{arXiv preprint arXiv:1806.03713}.

\bibitem[{Krizhevsky et~al.(2012)Krizhevsky, Sutskever, and
  Hinton}]{krizhevsky2012imagenet}
Alex Krizhevsky, Ilya Sutskever, and Geoffrey~E Hinton. 2012.
\newblock Imagenet classification with deep convolutional neural networks.
\newblock In \emph{Advances in neural information processing systems}, pages
  1097--1105.

\bibitem[{Kuhn(1955)}]{kuhn55}
Harold~W. Kuhn. 1955.
\newblock The {Hungarian Method} for the assignment problem.
\newblock \emph{Naval Research Logistics Quarterly}, page 83–97.

\bibitem[{Maddison et~al.(2014)Maddison, Tarlow, and
  Minka}]{maddision2014gumbletrick}
Chris~J Maddison, Daniel Tarlow, and Tom Minka. 2014.
\newblock \href {http://papers.nips.cc/paper/5449-a-sampling.pdf} {A*
  sampling}.
\newblock In \emph{Advances in Neural Information Processing Systems 27}, pages
  3086--3094.

\bibitem[{Nakashole and Mitchell(2014)}]{nakashole2014language}
Ndapandula Nakashole and Tom~M Mitchell. 2014.
\newblock Language-aware truth assessment of fact candidates.
\newblock In \emph{ACL}.

\bibitem[{Ojha and Talukdar(2017)}]{KGEval}
Prakhar Ojha and Partha Talukdar. 2017.
\newblock {KGEval}: Accuracy estimation of automatically constructed knowledge
  graphs.
\newblock In \emph{EMNLP}.

\bibitem[{Over and Yen(2004)}]{duc2004}
Paul Over and James Yen. 2004.
\newblock Introduction to {DUC}-2004: An intrinsic evaluation of generic news
  text summarization systems.
\newblock In \emph{Document Understanding Workshop}.

\bibitem[{Padia et~al.(2018{\natexlab{a}})Padia, Ferraro, and
  Finin}]{padiaFever2018}
Ankur Padia, Francis Ferraro, and Tim Finin. 2018{\natexlab{a}}.
\newblock {{UMBC-FEVER}: Claim verification using Semantic Lexical Resources}.
\newblock In \emph{First Workshop on Fact Extraction and Verification (EMNLP)}.

\bibitem[{Padia et~al.(2018{\natexlab{b}})Padia, Ferraro, and
  Finin}]{padia2018kgcleaner}
Ankur Padia, Frank Ferraro, and Tim Finin. 2018{\natexlab{b}}.
\newblock {KGCleaner}: Identifying and correcting errors produced by
  information extraction systems.
\newblock \emph{arXiv preprint arXiv:1808.04816}.

\bibitem[{Parikh et~al.(2016{\natexlab{a}})Parikh, T{\"a}ckstr{\"o}m, Das, and
  Uszkoreit}]{parikh16}
Ankur Parikh, Oscar T{\"a}ckstr{\"o}m, Dipanjan Das, and Jakob Uszkoreit.
  2016{\natexlab{a}}.
\newblock \href {https://doi.org/10.18653/v1/D16-1244} {A decomposable
  attention model for natural language inference}.
\newblock In \emph{EMNLP}, pages 2249--2255. ACL.

\bibitem[{Parikh et~al.(2016{\natexlab{b}})Parikh, T{\"a}ckstr{\"o}m, Das, and
  Uszkoreit}]{parikh2016decomposable}
Ankur~P Parikh, Oscar T{\"a}ckstr{\"o}m, Dipanjan Das, and Jakob Uszkoreit.
  2016{\natexlab{b}}.
\newblock A decomposable attention model for natural language inference.
\newblock \emph{arXiv preprint arXiv:1606.01933}.

\bibitem[{Pennington et~al.(2014)Pennington, Socher, and
  Manning}]{pennington2014glove}
Jeffrey Pennington, Richard Socher, and Christopher Manning. 2014.
\newblock Glove: Global vectors for word representation.
\newblock In \emph{EMNLP}.

\bibitem[{P{\'e}rez-Rosas et~al.(2017)P{\'e}rez-Rosas, Kleinberg, Lefevre, and
  Mihalcea}]{perez2017automatic}
Ver{\'o}nica P{\'e}rez-Rosas, Bennett Kleinberg, Alexandra Lefevre, and Rada
  Mihalcea. 2017.
\newblock Automatic detection of fake news.
\newblock \emph{arXiv preprint arXiv:1708.07104}.

\bibitem[{Pujara et~al.(2013)Pujara, Miao, Getoor, and
  Cohen}]{pujara2013knowledge}
J.~Pujara, H.~Miao, L.~Getoor, and W.~Cohen. 2013.
\newblock Knowledge graph identification.
\newblock In \emph{Int. Semantic Web Conf.}

\bibitem[{{Qi Zeng, Quan Zhou, and Shanshan Xu}(2018)}]{zengNeuralStance}
{Qi Zeng, Quan Zhou, and Shanshan Xu}. 2018.
\newblock {Neural Stance Detectors for Fake News Challenge}.
\newblock http://bit.ly/StanFNC.

\bibitem[{Richardson and Domingos(2006)}]{richardson2006markov}
Matthew Richardson and Pedro Domingos. 2006.
\newblock Markov logic networks.
\newblock \emph{Machine learning}, 62(1-2):107--136.

\bibitem[{Riedel et~al.(2017)Riedel, Augenstein, Spithourakis, and
  Riedel}]{riedel2017simple}
Benjamin Riedel, Isabelle Augenstein, Georgios~P Spithourakis, and Sebastian
  Riedel. 2017.
\newblock A simple but tough-to-beat baseline for the fake news challenge
  stance detection task.
\newblock \emph{arXiv preprint arXiv:1707.03264}.

\bibitem[{Samadi et~al.(2016)Samadi, Talukdar, Veloso, and
  Blum}]{samadi2016claimeval}
Mehdi Samadi, Partha~Pratim Talukdar, Manuela~M Veloso, and Manuel Blum. 2016.
\newblock {ClaimEval}: Integrated and flexible framework for claim evaluation
  using credibility of sources.
\newblock In \emph{AAAI}.

\bibitem[{Samadi et~al.(2013)Samadi, Veloso, and Blum}]{samadi2013openeval}
Mehdi Samadi, Manuela~M Veloso, and Manuel Blum. 2013.
\newblock {OpenEval}: Web information query evaluation.
\newblock In \emph{AAAI}.

\bibitem[{{Sean Bird, Doug Sibley, and Yuzi Pan}(2017)}]{talosFNC}
{Sean Bird, Doug Sibley, and Yuzi Pan}. 2017.
\newblock {Talos Targets Disinformation with Fake News Challenge Victory}.
\newblock http://bit.ly/TalosFNC.

\bibitem[{Sundheim(1992)}]{sundheim1992muc4}
Beth Sundheim. 1992.
\newblock Proceedings of the fourth message understanding conference ({MUC}-4).

\bibitem[{Thorne et~al.(2018)Thorne, Vlachos, Christodoulopoulos, and
  Mittal}]{thorne2018fever}
James Thorne, Andreas Vlachos, Christos Christodoulopoulos, and Arpit Mittal.
  2018.
\newblock {FEVER}: a large-scale dataset for fact extraction and verification.
\newblock \emph{arXiv preprint arXiv:1803.05355}.

\bibitem[{Vosoughi et~al.(2018)Vosoughi, Roy, and Aral}]{vosoughi2018spread}
Soroush Vosoughi, Deb Roy, and Sinan Aral. 2018.
\newblock The spread of true and false news online.
\newblock \emph{Science}, 359(6380):1146--1151.

\bibitem[{Yu et~al.(2014)Yu, Huang, Cassidy, Ji, Wang, Zhi, Han, Voss, and
  Magdon-Ismail}]{yu2014wisdom}
Dian Yu, Hongzhao Huang, Taylor Cassidy, Heng Ji, Chi Wang, Shi Zhi, Jiawei
  Han, Clare Voss, and Malik Magdon-Ismail. 2014.
\newblock The wisdom of minority: Unsupervised slot filling validation based on
  multi-dimensional truth-finding.
\newblock In \emph{COLING}.

\end{thebibliography}
\bibliographystyle{acl_natbib}

\end{document}